\documentclass[conference]{IEEEtran}
\IEEEoverridecommandlockouts

\usepackage{cite}
\usepackage{amsmath,amssymb,amsfonts}
\usepackage{algorithm}
\usepackage{algorithmic}
\usepackage{graphicx}
\usepackage{textcomp}
\usepackage{xcolor}
\usepackage{amsthm}
\usepackage{adjustbox}
\usepackage{booktabs}
\usepackage{multirow}
\usepackage{import}
\usepackage{url}
\usepackage{comment}

\usepackage{enumerate}

\usepackage{enumitem}

\def\BibTeX{{\rm B\kern-.05em{\sc i\kern-.025em b}\kern-.08em
    T\kern-.1667em\lower.7ex\hbox{E}\kern-.125emX}}
\makeatletter
\newcommand{\linebreakand}{%
  \end{@IEEEauthorhalign}
  \hfill\mbox{}\par
  \mbox{}\hfill\begin{@IEEEauthorhalign}
}
\makeatother

\begin{document}

\theoremstyle{definition}
\newtheorem{problem}{Problem}

\newtheorem{definition}{Definition}[section]

\title{{\it Few Labels are all you need:} \\ A Weakly Supervised Framework for Appliance Localization in Smart-Meter Series}

\author{\IEEEauthorblockN{Adrien Petralia}
\IEEEauthorblockA{\textit{EDF R\&D - Université Paris Cité} \\
adrien.petralia@gmail.com}
\and
\IEEEauthorblockN{Paul Boniol}
\IEEEauthorblockA{\textit{Inria, ENS, PSL, CNRS} \\
paul.boniol@inria.fr}
\and
\IEEEauthorblockN{Philippe Charpentier}
\IEEEauthorblockA{\textit{EDF R\&D} \\
philippe.charpentier@edf.fr}
\and
\IEEEauthorblockN{Themis Palpanas}
\IEEEauthorblockA{\textit{Université Paris Cité - IUF} \\
themis@mi.parisdescartes.fr}
}

\maketitle

\begin{abstract}
Improving smart grid system management is crucial in the fight against climate change, and enabling consumers to play an active role in this effort is a significant challenge for electricity suppliers. 
In this regard, millions of smart meters have been deployed worldwide in the last decade, recording the main electricity power consumed in individual households.
This data produces valuable information that can help them reduce their electricity footprint; nevertheless, the collected signal aggregates the consumption of the different appliances running simultaneously in the house, making it difficult to apprehend.
Non-Intrusive Load Monitoring (NILM) refers to the challenge of estimating the power consumption, pattern, or on/off state activation of individual appliances using the main smart meter signal.
Recent methods proposed to tackle this task are based on a fully supervised deep-learning approach that requires both the aggregate signal and the ground truth of individual appliance power.
However, such labels are expensive to collect and extremely scarce in practice, as they require conducting intrusive surveys in households to monitor each appliance.
In this paper, we introduce CamAL, a weakly supervised approach for appliance pattern localization that only requires information on the presence of an appliance in a household to be trained.
CamAL merges an ensemble of deep-learning classifiers combined with an explainable classification method to be able to localize appliance patterns.
Our experimental evaluation, conducted on 4 real-world datasets, demonstrates that CamAL significantly outperforms existing weakly supervised baselines and that current SotA fully supervised NILM approaches require significantly more labels to reach CamAL performances.  
The source of our experiments is available at: \url{https://github.com/adrienpetralia/CamAL}.
This paper appeared in ICDE 2025.
\end{abstract}


\section{Introduction}

Managing electricity consumption at the individual level has become a critical challenge in achieving more efficient smart grid management and contributing to the global effort to reduce energy usage. 
In response, energy suppliers such as EDF (Electricité De France) have begun offering various services to help customers better understand and manage their electricity consumption. 
Energy suppliers are installing meters that record the total aggregate electricity power consumed in the household at regular intervals.
Although this information provides valuable data that suppliers already use for diverse applications such as forecasting energy demand, there is a need to develop solutions to extract detailed information on household consumption. 
Indeed, the collected signal results from the addition of all the appliances operating simultaneously, and suppliers face the challenge of extracting detailed information from these data, such as \emph{if} and \emph{when} a specific appliance has been used looking only at the aggregated consumption time series.
Although extracting this valuable information is challenging due to the complexity of the aggregated data, it is crucial to help consumers manage their electricity consumption and provide them with more insight into their usage.

\begin{figure}[tb]
    \centering
    \includegraphics[width=0.98\linewidth]{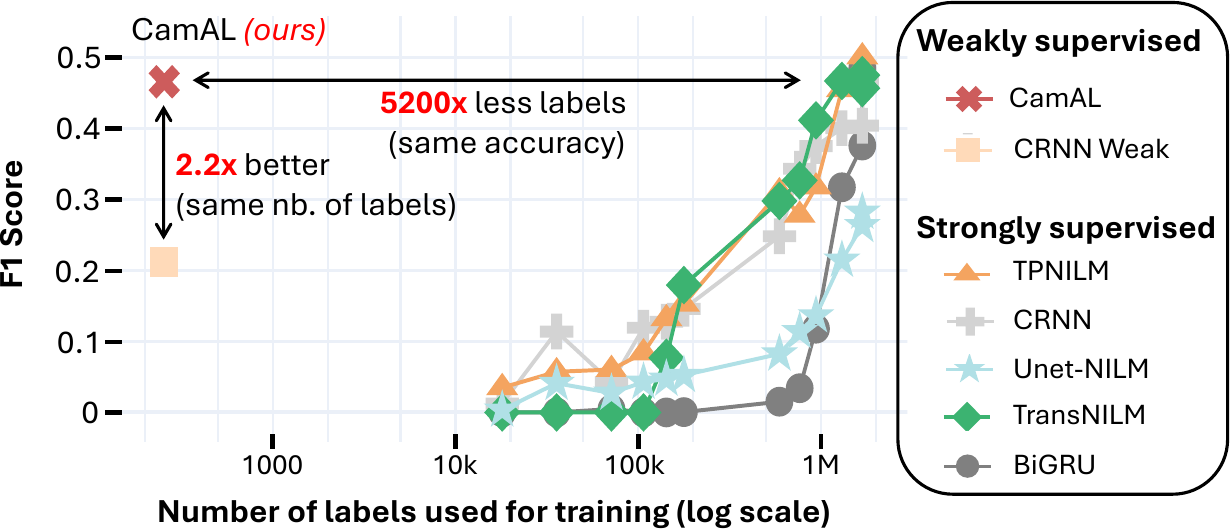}
    \caption{Localization accuracy versus number of training labels for CamAL compared to six baseline methods on the dishwasher case from the IDEAL dataset. CamAL and the weakly supervised baseline are trained using only one label per house, indicating appliance ownership.
    }
    \vspace{-0.5cm}
    \label{fig:introfig}
\end{figure}

Non-Intrusive Load Monitoring (NILM) refers to the challenge of estimating the power consumption, usage patterns, or on/off state activation of individual appliances using only aggregate power readings from a household. 
Early NILM solutions approached this task as an optimization problem, relying on Combinatorial Optimization (CO) to estimate the proportion of total power consumed by active appliances at each time step~\cite{hart_nilm_1992}.
Over the past decade, NILM research has surged, fueled by the release of publicly available smart meter datasets~\cite{Kolter2011REDDA, ukdale, refitdataset}.  
These datasets provide both aggregate power readings and appliance-level power consumption, often referred to as \emph{strong labels}, which indicate the exact activation state and consumption power of individual appliances for each timestamp. 
Consequently, NILM approaches predominantly rely on fully supervised, primarily deep-learning models that require extensive strong-label data for training (cf. \emph{strongly supervised} approaches in Figure~\ref{fig:introfig}). 

While these datasets have enabled the development and benchmarking of new algorithms, they are limited—typically encompassing data from only about a dozen households—and do not represent the diversity of appliances owned by all consumers. 
As a result, electricity suppliers need to invest in collecting their own data by instrumenting a large number of households.
However, conducting such a survey is expensive in terms of time, and $\text{CO}_2$ emissions, as it requires sending technicians to individual households to install sensors that measure the consumption of each appliance.
At the same time, it is also very expensive in terms of money: collecting ground truth appliance-level data from a mere few dozen households costs several hundreds of thousands of euros.

In practice, electricity suppliers 
only have at their disposal
the information of an appliance's activation (or not) within a time frame, meaning one label for an entire series, so-called \emph{weak labels}.
Particularly, in challenging but more realistic scenarios, they only know the presence of the appliances in the household, meaning one label for an entire long series, without any guarantee on \emph{when} the appliance is effectively used.
Unfortunately, recent proposed NILM approaches cannot be trained and operate with such scarce labels: trying to train a NILM solution with only one label for an entire series (e.g., by replicating the label for all time steps) implies that it can no longer be used to localize an appliance; indeed, NILM solutions provide a probability of detection for each timestamp to be able to localize it.
To address this issue, a recent study introduced the appliance localization problem using weak labels~\cite{weaknilm}. 
The authors proposed a deep-learning approach that formulates the challenge as a Multiple Instance Learning (MIL) problem. 
In this framework, the model is trained using only one label per series or by combining both strong and weak labels when available.
However, their results on real-world challenging datasets showed that accuracy is notably low when using only weak labels; they had to combine both strong and weak labels to achieve better performance.
Moreover, their approach was tested only on datasets that provide individual power consumption data for each appliance. 
It has not been evaluated in realistic scenarios where only the possession of the appliance in the household is known, without any data on when the appliance is actually used.

Recent studies have been conducted to detect the presence of appliances in consumption series using weakly supervised approaches~\cite{dengetal, eEnergy_ApplDetection, VLDB_TransApp}.
In these studies, the appliance detection problem is cast as a time series classification problem, in which a classifier is trained using only one label per electrical time series, i.e., we only know whether the appliance has been switched ON in a given period.
Although these methods show promising results in detecting \emph{if} an appliance has been used, they cannot determine \emph{when} the device has been switched on.
At the same time, a few recent works have shown that classification-based explainability methods can be used to understand a classifier's decision by identifying the part of a time series that contributed to the label prediction~\cite{lime, shape, cam, gradcam, dcam}.
These approaches have been tested on time series for explainable classification and anomaly detection tasks~\cite{ResNet_tsc_2017, Boniol_Meftah_Remy_Didier_Palpanas_2023} with promising results but have never been used for appliance localization.
For the NILM problem, such methods could enable the localization of appliances while being trained using labels that only indicate \emph{if} a device is turned on during a large time frame, significantly reducing the number of needed labels that current NILM approaches require.
In addition, several studies~\cite{snorkelvldb, snorkelsigmod} demonstrate the importance of leveraging weak labels to efficiently solve diverse real-world problems while using a few amounts of strong labels.

While {\it \textbf{few labels are all we have}}, we demonstrate in this paper that {\it \textbf{few labels are all we need}}.
This paper investigates for the first time the combination of explainable classification approaches to tackle the appliance pattern localization problem.
Overall, our framework, called CamAL, contains the following steps: 
(i) We train a set of convolutional neural network classifiers on smart meter consumption series, using labels that indicate whether a specific appliance was turned on within a large given time frame.
(ii) After training, our method performs localization by extracting and aggregating the class activation maps (CAM) from the classifiers if the appliance is detected.
(iii) Finally, we post-process the aggregated CAM to refine the prediction and output the most probable timestamps corresponding to the appliance usage.

We empirically compare CamAL to current state-of-the-art NILM methods for appliance localization across multiple real-world datasets, showing that CamAL often achieves comparable or superior performance while requiring up to three orders of magnitude fewer labels (see Figure~\ref{fig:introfig}).
More specifically, CamAL significantly outperforms existing weakly supervised approaches and strongly supervised NILM methods trained with the same number of labels—achieving up to a twofold improvement (see Figure~\ref{fig:introfig})—and scales more efficiently to large datasets than any of these methods.

Overall, we demonstrate that our approach is more appropriate than classical NILM methods for use cases without access to per-timestamp labels, which corresponds to the vast majority of realistic and industrial applications.
Consequently, CamAL is the first real non-invasive load monitoring method as it does not require practitioners to physically enter each household to install per-appliance sensors to train a solution. 
In addition, we demonstrate that CamAL scales to the real-world (possession only) datasets currently available to suppliers thanks to its weakly supervised paradigm, making it much faster to train than current NILM solutions.
Even though the inference time of CamAL is not significantly faster than existing NILM methods, it drastically surpasses the weakly supervised one and lets practitioner saves a significant amount of time, money, cost of storage, and CO$_2$ emissions necessary to create the labels to train a strongly supervised method.


Overall, our contributions are the following:

\begin{itemize}[noitemsep, topsep=0pt, parsep=0pt, partopsep=0pt, leftmargin=0.4cm]
        \item We first formalize the problem of appliance detection and localization, and we discuss the proposed methods available in the literature to solve these problems (\textbf{Section~\ref{sec:backprel}}).
        \item We propose CamAL (\textit{Class Activation Map based Appliance Localization}) a novel weakly supervised method for appliance pattern localization based on time series classification and explainable AI that requires only the appliance's possession label for training (\textbf{Section~\ref{sec:proposedappraoch}}).
        \item We experimentally compare CamAL with state-of-the-art NILM approaches for appliance localization, and we empirically demonstrate that the NILM problem can be solved with significantly fewer labels (\textbf{Section~\ref{sec:experiments}}). 
        \item We demonstrate through a public dataset and a real industrial use case that CamAL can operate with one label only (i.e., the possession or not of the appliance), making our model the first accurate and scalable real Non-Intrusive Load monitoring system (\textbf{Section~\ref{sec:application}}).
        \item We conclude by showing that our generated weak labels can be used to compensate for the scarcity or strong labels to maintain a high accuracy of strongly supervised NILM approaches (\textbf{Section~\ref{sec:augmentation}}).
\end{itemize}


\section{Background and Related Work}
\label{sec:backprel}

A smart meter signal is a univariate time series $\boldsymbol{x} = \bigl(\mathbf{x}_{t_1},\mathbf{x}_{t_2},\,\ldots,\mathbf{x}_{t_T}\bigr)$ of $T$ timestamped power consumption readings. 
The meter reading is defined as the time difference $\Delta_t = t_{i} - t_{i-1}$ between two consecutive timestamps $t_i$. 
Each element $\boldsymbol{x}_t$ (in Watts or Watt-hours) represents either the actual power at time $t$ or the average power over the interval $\Delta_{t_i}$.

\begin{figure}[tb]
    \centering
    \includegraphics[width=\linewidth]{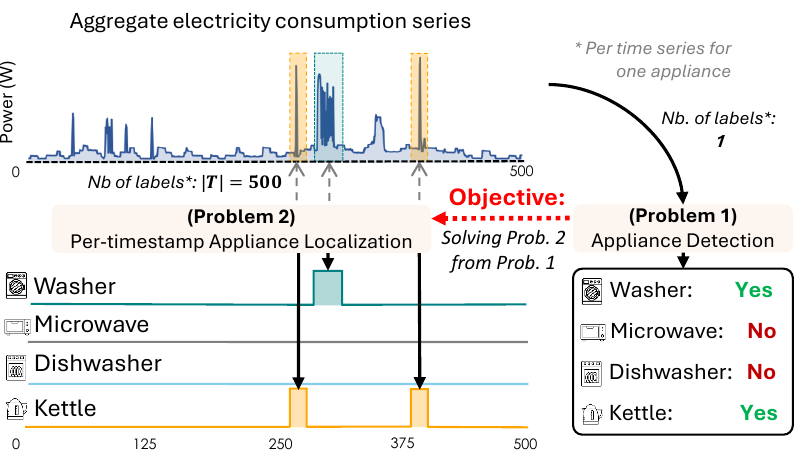}
    \caption{Illustration of the appliance detection (1 label needed) and per-timestamp appliance localization ($|T|$ labels needed) problems. Our objective in this paper is to solve Problem 2 from Problem 1.}
    \label{fig:problem-def}
    \vspace*{-0.3cm}
\end{figure}

The aggregate power consumption is defined as the sum of $N$ appliance power signals $a_1(t), a_2(t), \ldots, a_N(t)$ that run simultaneously plus some noise $\epsilon(t)$, accounting for measurement errors. 
Formally, it is defined as:
{\small
\begin{equation}
\label{eq:problem1}
    x(t)=\sum^{N}_{j=1} a_j(t) + \epsilon(t)
\end{equation}
}
where $x(t)$ is the total power consumption measured by the main meter at timestep $t$; $N$ is the total number of appliances connected to the smart meter; and $\epsilon(t)$ is defined as the noise or the measurement error at timestep $t$.

Practitioners are interested in solving two problems: (i) discovering {\it if} an appliance has been activated (Appliance Detection Problem), and (ii) identifying {\it when} an appliance has been used (Per-Timestamp Appliance Localization Problem). The two problems are formalized as follows: 

\begin{problem}[Appliance Detection (cf. Figure~\ref{fig:problem-def})]
\label{prob:detection}
Given an aggregate consumption smart meter series $\boldsymbol{x} = \bigl(\mathbf{x}_{t_1},\mathbf{x}_{t_2},\,\ldots,\mathbf{x}_{t_T}\bigr) \in \mathbb{R}^+_{T}$, an appliance $a$, we want to know if $a$ has been used in $\boldsymbol{x}$ (i.e., was in an "ON" state, regardless of the time and number of activations). 
\end{problem}



\begin{problem}[Per-timestamp Appliance Localizaton (cf. Figure~\ref{fig:problem-def})]
\label{prob:localisation}
The total active power consumed in a household is denoted by $x(t)$, the active power of the $j$-th appliance by $a_j(t)$, and its state by $s_j(t) \in \{0, 1\}$. 
Then we have:
{\small
\begin{equation}
x(t) = \sum_{j=1}^{N} s_j(t)a_j(t) + \varepsilon(t), 
\end{equation}
} 
where $\varepsilon(t)$ represents the measurement noise, and
{\small
\begin{equation}
s_j(t) = 
\begin{cases} 
0, & \text{if appliance } j \text{ is OFF at time index } t,\\
1, & \text{if appliance } j \text{ is ON at time index } t.
\end{cases}
\end{equation}
} 
We want to compute the consumption (or activation) of appliance $j$, $a_j(t)$, from $x(t)$. 
\end{problem}

In order to solve Problem~\ref{prob:localisation}, we can rewrite Equation~\ref{eq:problem1} as: 
\begin{equation}
x(t) = s(t)a(t) + v(t),
\end{equation}
where the first term is the power of the appliance of interest, and $v(t)$ is a cumulative noise term corresponding to the sum of all the other appliances running simultaneously.

In cases where the objective is the direct estimation of the individual active power signal $a(t)$, NILM is treated as a regression problem and has been approached either as a denoising task or as a blind source separation task~\cite{themis_reviewnilm2024}. 
Conversely, when the objective is to estimate the appliance state $s(t)$, NILM represents a classification problem~\cite{themis_reviewnilm2024}. 
In both cases, the algorithm utilizes only the knowledge of the aggregate signal $x(t)$.
This work focuses on appliance status detection, aiming to estimate the state variables $s(t)$ of the appliance of interest.

Note that the proposed methods to solve Problem~\ref{prob:localisation} require one label per timestamp and per appliance. 
On the contrary, methods aiming to solve Problem~\ref{prob:detection}, i.e. time series classifiers, require only labels indicating if an appliance has been used within a time frame.
Such labels are significantly easier to collect with non-intrusive solutions, such as asking people to answer questionnaires.
In the following sections, we discuss the existing methods proposed in the literature for solving the two different problems.

\subsection{Non-Intrusive Load Monitoring}

In the literature, Problem~\ref{prob:localisation} and the corresponding proposed methods are usually referred to as energy disaggregation~\cite{hart_nilm_1992} (a.k.a. NILM).
NILM relies on identifying the power consumption (or on/off state activation) of individual appliances using only the total aggregated load curve~\cite{themis_reviewnilm2024, eEnergy_ApplDetection, eEnergy_laviron_activation, reviewnilm}.

\subsubsection{Sequence-to-Sequence Approaches}

Most methods proposed to solve Problems~\ref{prob:localisation} are sequence-to-sequence approaches. 
Early NILM solutions involved Combinatorial Optimization (CO) to estimate the proportion of total power consumption used by distinct active appliances at each time step~\cite{hart_nilm_1992}.
Later, semi-supervised and unsupervised machine learning algorithms, such as factorial hidden Markov Models (FHMM), were investigated~\cite{FHMM_2011}.
These solutions mainly used expert domain knowledge, and the accuracy reported is low compared to supervised ones~\cite{themis_reviewnilm2024}.
NILM gained popularity in the late 2010s, following the release of smart meter datasets~\cite{Kolter2011REDDA, ukdale, refitdataset} allowing training and benchmarking supervised methods that demonstrate significantly better performance than semi-supervised and unsupervised ones~\cite{themis_reviewnilm2024}.
Kelly et al.~\cite{Kelly_2015} were the first to investigate deep learning (DL) approaches to tackle the NILM problem and assessed the superiority of three different DL architectures against FHMM and CO.
Since then, numerous studies have proposed different DL methods for solving the NILM based on various kinds of architectures.
A plain Fully Convolutional Network (FCN) was proposed~\cite{cnn_zhang_2018}, followed by a Dilated Attention ResNet~\cite{daresnet_2019}, and, inspired by solutions for image segmentation~\cite{unet_2015}, the Temporal Pooling~\cite{tpnilm_2020} and UNet~\cite{unet_nilm_2020} architectures were adapted to detect appliance status activation.
More recently, hybrid architectures were investigated, such as BiGRU~\cite{BiGRU_2023} that mixes Convolution layers and Bidirectional Gated Recurrent Units and inspired by Transformer-based approaches such as BERT~\cite{bert_2019}, BERT4NILM~\cite{bert4nilm_2020} was proposed, followed by TransNILM~\cite{transnilm_2022}, an extension of~\cite{tpnilm_2020} specifically designed to solve the appliance localization problem.

Nevertheless, all the aforementioned sequence-to-sequence methods require the individual appliance load curves to be trained.
Gathering such labels requires installing sensors measuring each device in many different houses. 
Generating large enough datasets to train accurate sequence-to-sequence methods is costly, invasive, and time-consuming.

\subsubsection{Weakly Labeled Approaches}

A recent study~\cite{weaknilm} introduced a weak supervision paradigm for NILM, casting it as a Multiple-Instance Learning (MIL) problem. 
The authors proposed a Convolutional Recurrent Neural Network (CRNN) designed to utilize both weak labels (sequence-level annotations indicating whether an appliance was active within a segment) and strong labels (fine-grained, frame-level annotations).
The method required a combination of strong and weak labels to achieve acceptable accuracy. 
Furthermore, it was only tested on datasets with appliance-level power data, making it unsuitable for scenarios where only appliance possession information is available without usage details. 
These constraints limit its practicality and scalability in real-world applications.

\subsection{Appliance Detection}

Instead of tackling Problem~\ref{prob:localisation} directly (usually not applicable in practice due to the lack of labels), we can target Problem~\ref{prob:detection}, which can be seen as an intermediary step before Problem~\ref{prob:localisation}.
As mentioned earlier, Problem~\ref{prob:detection} consists of detecting which devices have been used within a large time frame (instead of a prediction per timestamp). 
Problem~\ref{prob:detection} can be treated as a supervised binary classification problem and solved using a trained times series classifier~\cite{eEnergy_ApplDetection}.
In contrast to sequence-to-sequence methods, such classifiers require labels indicating if an appliance has been used within a time frame. 
Such labels are significantly easier to collect with non-intrusive solutions such as surveys (questionnaires) sent to the customers.
Recent studies and benchmarks~\cite{dengetal, eEnergy_ApplDetection, VLDB_TransApp} evaluated the state-of-the-art time series classification methods on this problem and revealed that deep-learning methods are the most accurate and scalable to solve the appliance detection problem.
Nevertheless, the output of such classifiers only indicates if a device is turned on during a time frame but cannot indicate when exactly the appliance has been used (i.e., solving Problem~\ref{prob:localisation} directly).


\subsection{Explainable Artificial Intelligence}

Explainable artificial intelligence (XAI) is a growing field of interest. 
Using deep neural networks, common interpretable techniques employ approaches based on heatmap visualizations derived from the model architecture.
These visualization techniques can be categorized into three distinct groups: gradient-based attention visualizations~\cite{gradcam}, visualization based on Class Activation Maps (CAMs)~\cite{cam}, and perturbation-based input manipulation~\cite{10.1145/3447548.3467166, lime, shape}. 
Notably, CAM and gradient-based approaches are widely applied in computer vision for object localization~\cite{gradcam, gradcampp, scorecam}.
These methods have also been explored in the time series domain, particularly for time series classification, where explainability in this context aims to identify discriminative features that explain why a time series belongs to a specific class~\cite{ResNet_tsc_2017, shape}, as well as for anomaly detection~\cite{dcam,Boniol_Meftah_Remy_Didier_Palpanas_2023}.
Techniques such as LIME~\cite{lime}, WindowSHAP~\cite{shape}, and class-activation-based methods such as CAM~\cite{cam} and Grad-CAM~\cite{gradcam} have been adapted to this task.
However, none of these methods have been applied to address the Non-Intrusive Load Monitoring (NILM) problem. 
By applying an explanation method to a classifier trained to detect whether an appliance has been turned on, it is possible to localize the event by highlighting the most significant timestamps contributing to the classification decision. 
Therefore, using CAM applied on top of a trained classifier for appliance detection (solving Problem~\ref{prob:detection})  can be used to localize the appliance pattern in a smart meter consumption series (solving Problem~\ref{prob:localisation}). 
Formally: 

\begin{definition}[Class Activation Map]
For a given time series, and in the context of a trained deep-learning classifier that includes a Global Average Pooling (GAP) layer between the final convolutional layer and the last fully connected layer followed by a softmax activation; let denote \( f^k(t) \) the $k-th$ feature map at timestep $t$ of the last convolutional layer.
Therefore, for a class of interest $c$, the CAM explanation, written ${\text{CAM}}_c$, is defined as the weighted sum:
$\text{CAM}_c = \sum_k w^k_c f^k$
with $w_c^k$ representing the weights associated with class $c$ of the fully connected layer that functions as a classifier.
\end{definition}

\section{Research Questions}
\label{sec:researchquestions}

As a consequence, and as illustrated in red in Figure~\ref{fig:problem-def}, the objective of this paper is to use an explainability approach (such as CAM) applied on top of classifiers (trained for Problem~\ref{prob:detection}) to solve Problem~\ref{prob:localisation}.
This would mean that the NILM problem can be handled without conducting expensive surveys to gather labels to train accurate sequence-to-sequence solutions (i.e., requiring practitioners to install dedicated sensors per appliance in a large number of households).

From a business and industrial point of view, answering this question is crucial as using this approach would save a tremendous amount of time and money while reducing significantly CO$_2$ emissions.
We divide the latter into different research questions that we will address in this paper:

\begin{itemize}
    \item \textbf{RQ1}: \textit{Are weak labels enough to reach the performances of NILM methods trained on strong labels?}
    \item \textbf{RQ2}: \textit{Are Appliance Detection accuracy and Appliance Localization accuracy correlated?}
    \item \textbf{RQ3}: \textit{What are the optimal design choices to perform weakly supervised NILM?}
    \item \textbf{RQ4}: \textit{Is the information of appliance possession in this household (one label only) enough to train CamAL?}
    \item \textbf{RQ5}: \textit{Can we use CamAL predictions to train strongly supervised NILM approaches?}
\end{itemize}


\section{CamAL: A Weakly Supervised Approach for Appliance Pattern Localization}
\label{sec:proposedappraoch}

\begin{figure}[tb]
    \centering
    \includegraphics[width=0.85\linewidth]{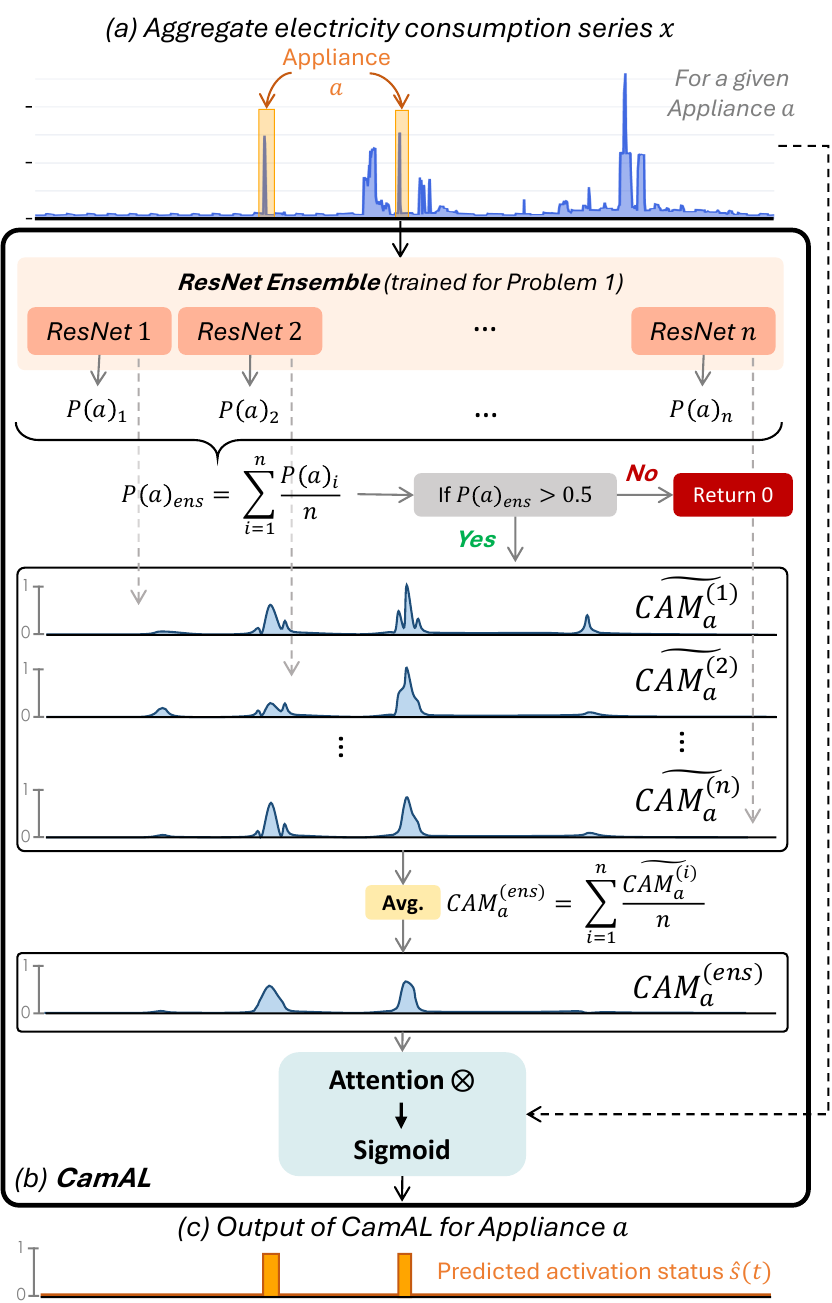}
    \vspace{-0.1cm}
    \caption{CamAL framework overview.}
    \vspace{-0.4cm}
    \label{fig:camalframework}
\end{figure}

We now describe CamAL, our proposed approach that enables the detection and localization of appliance patterns in aggregated consumption series.
CamAL can be decomposed into two parts (see Figure~\ref{fig:camalframework}): (1) an ensemble of deep-learning classifiers that performs the detection and (2) an explainability-based module that localizes the appliance (when detected).
The ensemble of deep learning classifiers is based on different Convolutional ResNet architectures with varying kernel sizes.
In simple terms, the explainability-based module can be described as extracting the CAM of all the different classifiers of the ensemble and using it as an attention mask to highlight the parts of the input sequence that contribute the most to the decision.
In this section, we first describe the ResNets ensemble used for appliance detection and then delve into the details of our appliance localization-based module.


\subsection{Step 1: An Ensembling Approach for Appliance Detection}
\label{subsec:appldetection}

Detecting whether an appliance has been used during a specified period can be framed as a time series classification (TSC) problem, as discussed in previous literature~\cite{eEnergy_ApplDetection}. 
For this task, a binary classifier is trained to recognize the use of an appliance, assigning a single label (0 or 1) to the entire time series. 
Previous research~\cite{eEnergy_ApplDetection, VLDB_TransApp} has shown that deep learning methods are particularly effective for this challenge. 
These approaches include convolution-based methods, such as non-deep learning ones such as Rocket and its variant~\cite{rocket, minirocket}  and deep learning ones, such as ResNet, InceptionTime~\cite{inceptiontime_fawaz}, as well as convolutional-transformer methods such as TransApp~\cite{VLDB_TransApp}. 
Nevertheless, non-deep learning convolutional-based methods like Rocket are not interpretable and, therefore, unsuitable for solving our problem.
Conversely, architectures such as InceptionTime and TransApp were designed as general, purpose models to achieve good classification performance regardless of the pattern length, in our case, different appliance usage patterns. 
In addition, these models are deeper and less efficient than simple ResNet architectures. 

Therefore, to develop an ensemble of classifiers that is both accurate and efficient, CamAL employs convolutional residual networks (ResNets) tuned for specific appliance patterns to detect the presence of an appliance in a given series. 
These networks are recognized for their accuracy, scalability, and well-studied decision-making processes on time series data, making them a robust backbone for our solution.

\subsubsection{CamAL ResNets Ensemble}

\begin{figure}[tb]
    \centering
    \includegraphics[width=0.8\linewidth]{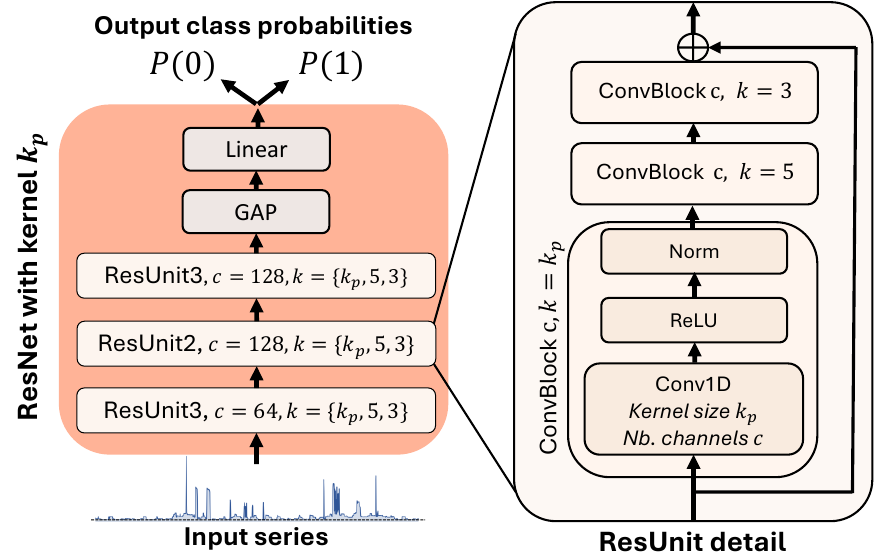}
    \vspace{-0.2cm}
    \caption{Detail of the ResNet architecture (for a specific kernel $k_p$).}
    \vspace{-0.5cm}
    \label{fig:resnet}
\end{figure}

The Residual Network (ResNet) architecture was introduced to address the gradient vanishing problem encountered in large CNNs~\cite{resnet_cv_2016}.
A ResNet architecture has been proposed for time series classification in~\cite{ResNet_tsc_2017} and has shown great performance on different benchmark~\cite{fawaz_tsc_review_2019}.
The architecture is composed of 3 stacked residual blocks connected by residual connections: this means that the input of a residual block is taken and added to its output. 
Each residual block comprises 3 convolutional blocks as described in the ConvNet architecture (same kernel size $\{8, 5, 3\}$, but each layer in a block uses the same number of filters).
The three residual blocks came with respectively $\{64, 128, 128\}$ filters, and, at this end, a global average pooling is performed along the temporal dimension followed by a linear layer and a softmax activation function to perform classification.

We leverage the proposed baseline~\cite{ResNet_tsc_2017} to an ensemble of $n$ networks differing in kernel sizes within the convolutional layers.
By default, we set $n=5$.
More specifically, the ensemble is based on an ensemble of networks trained with different kernel sizes $k_p$  (with $k_p \in K_p = \{5, 7, 9, 15, 25\}$). 
The ResNet architecture used in our ensemble is shown in Figure~\ref{fig:resnet} according to kernel size $k_p$.
This design choice is based on the intuition that varying the size of the kernel changes the receptive fields of the convolutional neural network (CNN), offering different levels of explainability.
We use the procedure described in Algorithm~\ref{algorithmcamal} to train the ResNet ensemble, which aims to train multiple ResNets with the same kernel $k_p$ and then select the networks that best detect the appliance $a$ regarding a validation dataset. 
We motivate the choice of the number of ResNets $n$ used in our ensemble as well as the introduction of different kernels $k_p$ in Section~\ref{sec:rq3}.

\begin{algorithm}[tb]
\caption{CamAL ResNet Ensemble Training for an Appliance $a$}
\label{algorithmcamal}
{\small
\begin{algorithmic}[1]
\REQUIRE training dataset $\mathcal{D}_{\text{train}}$, validation dataset $\mathcal{D}_{\text{validation}}$, number of ensemble networks $n$ (by default, $n=5$)
\STATE Split $\mathcal{D}_{\text{train}}$ into $\mathcal{D}_{\text{train-sub}}$ (80\%) and $\mathcal{D}_{\text{val-sub}}$ (20\%) to monitor training and prevent overfitting.
\FOR{each kernel size $k_p \in K_p$}
    \FOR{trial $t \in \{1, 2, 3\}$}
        \STATE Train a ResNet with kernel size $k_p$ on $\mathcal{D}_{\text{train-sub}}$.
        \STATE Evaluate the model on $\mathcal{D}_{\text{validation}}$ and store the validation loss.
    \ENDFOR
\ENDFOR
\STATE Collect all trained models and their validation losses.
\STATE Select the $n$ models with the lowest validation loss on $\mathcal{D}_{\text{validation}}$.
\RETURN Ensemble of $n$ models trained to detect $a$.
\end{algorithmic}
}
\end{algorithm}

\subsection{Step 2: Appliance Pattern Localization}
\label{subsec:explainability}

Identifying the discriminative features that influence a classifier's decision-making process has been extensively studied. 
Using deep-learning architecture for classification tasks, different methods have been proposed to highlight (i.e., localize) the parts of an input instance that contribute the most to the final decision of the classifier~\cite{cam, gradcam, dcam}.
Based on this previous work, we developed a specific CAM-based method to localize appliance patterns in a given consumption series.
Our approach involves extracting the CAMs from all the ResNets in the trained ensemble, computing their average, and applying the resulting map as an attention mask to the input series. 
This process highlights the regions in the time series that are most indicative of the appliance's operation while considering the shape of the aggregate signal to better localize the exact appliance activation time.
The detailed steps of our method are outlined below and depicted in Figure~\ref{fig:camalframework}:

\subsubsection{\textbf{ResNet Ensemble Prediction}} 
An aggregated input sequence $\boldsymbol{x}$ is fed into the ensemble of ResNet models. 
Each model predicts the probability of detection that appliance $a$ is present in $\boldsymbol{x}$. 
The ensemble prediction probability is computed by averaging the individual model probabilities:
$\text{Prob}_{\text{ens}} = \frac{1}{n} \sum_{i=1}^n \text{Prob}_i$,
where $n$ is the number of models in the ensemble, and $\text{Prob}_i$ is the prediction of model $i$.

\subsubsection{\textbf{Appliance Detection}} 
If the ensemble probability exceeds a threshold (e.g., $\text{Prob}_{\text{ens}} > 0.5$), the appliance is considered detected in the current window.
Otherwise, the appliance is undetected, and the activation status (i.e., localization) is set to 0 for each timestamp.

\subsubsection{\textbf{CAM Extraction}} 
If the appliance $a$ is detected, we extract each ResNet's CAM for class 1. 
As introduced before, for univariate time series, the CAM for class $c$ at timestamp $t$ is defined as:
$\text{CAM}^{(i)}_{c=1}(t) = \sum_{k} w^k_{c=1} \cdot f^k(t),$
where $w^k_c$ are the weights associated with the $k$-th filter for class $c$, and $f^k(t)$ is the activation of the $k$-th feature map at time $t$ for the CAM that correspond to the $i$-th ResNet in the ensemble.

\subsubsection{\textbf{CAM Processing}} 
Each $\text{CAM}^{(i)}$ is normalized to the range $[0, 1]$ by dividing it by its maximum value. 
Then, the average of each extracted CAM of the ensemble is computed as follows:
$\text{CAM}^{(ens)}(t) = \frac{1}{n} \sum_{i=1}^n \widetilde{\text{CAM}}^{(i)}(t).$

\subsubsection{\textbf{Attention Mechanism}}
$\text{CAM}^{(ens)}$ serves as an attention mask, highlighting the ensemble decision for each timestamp. 
We apply this mask to the input sequence through point-wise multiplication and pass the results through a sigmoid activation function to map the values in $[0,1]$: 
    $s(t) = \text{\emph{Sigmoid}}(\text{CAM}^{(ens)}(t) \circ \boldsymbol{x}(t)).$

\subsubsection{\textbf{Appliance Status}} 
The obtained signal is then rounded to obtain binary labels ($1$ if $s(t) \geq 0.5$), indicating the appliance's status and resulting in a binary time series $\hat{s}(t)$.

\subsection{From Binary Labels to Consumption per Appliance}
\label{sec:postprocessnilm}

To estimate the individual power consumption \(\hat{p}_a(t)\) of an appliance \(a\) using the predicted status signal \(\hat{s}(t)\) from CamAL, we employ a straightforward method. 
First, we multiply the binary status signal \(\hat{s}(t)\)—where \(\hat{s}(t) = 1\) by the mean power consumption \(P_a\) of the appliance (this parameter can be inferred from the dataset or provide by expert) as:

    $\hat{p}_a^\text{initial}(t) = \hat{s}(t) \cdot P_a$.

Then, to ensure that the estimated individual power consumption does not exceed the total aggregate power consumption at any given time \(t\), we apply a clipping operation that aims to adjust the estimated power so that it is always less than or equal to the observed aggregate consumption \(x(t)\):
$\hat{p}_a(t) = \min\left( \hat{p}_a^\text{initial}(t), \, x(t) \right)$.

\begin{table}[tb]
\caption{Datasets Details, Preprocessing Parameters and Appliance} 
\label{table:detaildatasetandparams}
\centering
\begin{adjustbox}{width=\linewidth,center}
\begin{tabular}{c|c|c|c|c|c}
    \toprule
    \textbf{Dataset} & \textbf{Nb. houses} & \textbf{Max. ffill} & \textbf{Appliance} & \textbf{ON Power} & \textbf{Avg. Power} \\
    \midrule
    \multirow{3}{*}{UKDALE} & \multirow{3}{*}{5} & \multirow{3}{*}{3 min} & Dishwasher & 300 W & 800 W \\
    & & & Microwave & 200 W & 1000 W \\
    & & & Kettle & 500 W & 2000 W \\
    \midrule
    \multirow{4}{*}{REFIT} & \multirow{4}{*}{20} & \multirow{4}{*}{3 min} & Dishwasher & 300 W & 800 W \\
    & & & Washing Machine & 300 W & 500 W \\
    & & & Microwave & 200 W & 1000 W \\
    & & & Kettle & 500 W & 2000 W \\
    \midrule
    \multirow{3}{*}{IDEAL} & 39  & \multirow{3}{*}{30 min} & Dishwasher & 300 W & 800 W \\
    & (+216 w/o & & Washing Machine & 300 W & 500 W \\
    & submeter) &  & Shower & 1000 W & 8000 W \\
    \midrule
    EDF EV & 24 & 1h30 & Electric Vehicle & 1000 W & 4000 W  \\
    \midrule
    \multirow{2}{*}{EDF Weak} & 558 (w/o  & \multirow{2}{*}{1h30} &  Electric Vehicle & \multirow{2}{*}{\slash} & \multirow{2}{*}{\slash}  \\
    & submeter) &  & (Possession only) & &   \\
    \bottomrule
\end{tabular}
\end{adjustbox}
\end{table}

\section{Experimental Evaluation}
\label{sec:experiments}

All experiments are performed on a 
server with 2 Intel Xeon Platinum 8260 CPUs, 384GB RAM, and 4 NVidia Tesla V100 GPUs with 32GB RAM.
The code (Python 3.12) is publicly available~\cite{camalcode}, as well as a corresponding demo~\cite{2025_petralia_devicescope_icde}. 

\subsection{Datasets}
\label{sec:datasets}

We use 5 datasets in our study. 
The first four—\textbf{UK-DALE}~\cite{ukdale}, \textbf{REFIT}~\cite{refitdataset}, and \textbf{IDEAL}~\cite{ideal} (all publicly available), and the private \textbf{EDF EV} dataset, provide both aggregate household power and individual appliance measurements. 
The fifth, EDF \textbf{EDF Weak}, is a private survey-based dataset containing only aggregate consumption data and electric vehicle ownership information. 
Further details are provided below.

\subsubsection{Public Datasets} UKDALE~\cite{ukdale} and REFIT~\cite{refitdataset} are two well-known 
datasets used in many research papers to assess the performance of NILM approaches~\cite{eEnergy_ApplDetection, cnn_zhang_2018, bert4nilm_2020, transnilm_2022}.
The two datasets contain high-frequency sampled data collected from small groups of houses in the UK and focus on small appliances. 
The IDEAL~\cite{ideal} dataset comprises data from 255 households in the United Kingdom.
For all the houses, the aggregate main power consumption of the house was recorded, and each participant filled out a questionnaire to provide some information about their household, including the type and number of appliances owned.
In addition, more detailed data are available for a subset of 39 households, including the individual electricity power consumption for different monitored appliances.

\subsubsection{EDF Datasets}
We use two private EDF datasets:

\noindent\textbf{[EDF EV]:} Data from 24 French households (July 2022--February 2024) with an average recording duration of 397 days 
(range: 175--587 days). Each house has 30-minute aggregate power readings and corresponding EV charger load curves.

\noindent\textbf{[EDF Weak]:} A survey-based dataset of 558 French households (September 2020--December 2022). 
Only total household power consumption was recorded, while EV ownership information was obtained via questionnaires.

\subsection{Data Processing}
\label{sec:dataprocessing}

According to the parameters reported in Table~\ref{table:detaildatasetandparams}, we resample and readjust recorded values to round timestamps by averaging the power consumed during the interval $\Delta_t$ and forward-filling the missing values. 

To meet a challenging real-world scenario in this study, we evaluate the model's performance using unseen data from different houses within the same dataset~\cite{themis_reviewnilm2024}. 
This means that distinct houses were used for training and evaluation to ensure robust performance assessment. 
In addition, we note that we select for each dataset the appliances that consumed the most power and are suitable for localization (in contrast to always ON devices such as the Fridge).

For the UKDALE dataset containing only 5 houses, we use houses 1, 3, and 4 for training while randomly selecting houses 2 and 5 as validation or test sets.
For all the other datasets, the houses used for train, valid, and test are randomly chosen. More precisely, the test set contains 2, 6, and 4 houses, and the validation set contains 2, 2, and 4 houses for REFIT, IDEAL, and EDF EV, respectively.


As two comparative baselines require the use of a non-overlapping window length $w=510$ as input~\cite{tpnilm_2020, transnilm_2022}, 
we slice the consumption data into non-overlapping subsequences of length $w=510$ for training and evaluating all models.
Subsequences containing any remaining missing values after our preprocessing are discarded. 
We note that we scaled the data by dividing the aggregate input consumption series by 1000 to ensure training stability.
The ground true status is calculated according to the "ON" status threshold reported in Table~\ref{table:detaildatasetandparams}.
In addition, before evaluating the models, we apply the process described in Section~\ref{sec:postprocessnilm} to all the models according to the appliance average power (i.e., $P_a$) reported in Table~\ref{table:detaildatasetandparams}. 

\subsection{Selected Baselines}
\label{sec:evalpipeline}

We compare our solution against different sequence-to-sequence strongly supervised baselines.
Our selection of competitors is based on their performance in previous studies~\cite{BiGRU_2023, transnilm_2022}).
We include two CNN-based architectures, \textbf{Unet-NILM}~\cite{unet_2015}, a UNet convolutional-based architecture, and \textbf{TPNILM}~\cite{tpnilm_2020}, a temporal pooling-based architecture.
In addition, we include a recently proposed recurrent-based architecture, \textbf{BiGRU}~\cite{BiGRU_2023}, that combines Convolution and Recurrent layers and \textbf{TransNILM}~\cite{transnilm_2022}, a SotA Tranformer-based architecture based on temporal pooling.

Finally, we include the CRNN architecture proposed in~\cite{weaknilm} that we decline in two versions.
In the rest of our paper, we refer to \textbf{CRNN} for the supervised version of the architecture trained using both strong and weak labels for each subsequence. 
Conversely, we refer to \textbf{CRNN Weak} for the weakly supervised version of the architecture trained only weak labels. 

For training, the strongly supervised baseline received one label per timestamp, while the two weakly supervised ones received one label per subsequence.
For all models, we used the default parameters provided by the authors and trained them using the Binary Cross Entropy Loss. 
In addition, we note that each baseline, including CamAL, is trained in one model per appliance setting.

\noindent \textbf{[Theoretical Model Complexity]}
We derive the theoretical complexity and the number of trainable parameters of selected baselines, including CamAL (see Table~\ref{tab:complexity-params}).
We use $L$ for the time series length; $C$, $K$ are the number of channels and the kernel size used in a convolutional layers kernel, respectively; $I$, $H$ are the input hidden dimensions and the number of recurrent units used in a recurrent layer, respectively; $D$ is the inner dimension used in a Transformer layer. 

\begin{table}
    \caption{Theoretical complexity and number of trainable parameters for the different baselines.}
    \label{tab:complexity-params}
    \begin{adjustbox}{width=\columnwidth, center}
    \begin{tabular}{c|c|c}
    \toprule
    \textbf{Model} & \textbf{Theoretical Complexity} & \textbf{\# Trainable Param.} \\ 
    \midrule
    \textbf{CamAL} 
      & $O\bigl(n_{\text{ResNet}} \cdot L \cdot C^2 \cdot K\bigr)$ 
      & $n_{\text{ResNet}} \times 570\text{K}$ \\
    \textbf{CRNN (Weak/Strong)} 
      & $O\bigl(L \cdot C^2 \cdot K \cdot (I \cdot H + H^2)\bigr)$ 
      & $1049\text{K}$ \\
    \textbf{BiGRU}
      & $O\bigl(L \cdot C^2 \cdot K \cdot (I \cdot H + H^2)\bigr)$
      & $244\text{K}$ \\
    \textbf{Unet-NILM} 
      & $O\bigl(L \cdot C^2 \cdot K\bigr)$
      & $3197\text{K}$ \\
    \textbf{TPNILM}
      & $O\bigl(L \cdot C^2 \cdot K\bigr)$
      & $328\text{K}$ \\
    \textbf{TransNILM}
      & $O\bigl(L^2 \cdot D \cdot L \cdot C^2 \cdot K \cdot (I \cdot H + H^2)\bigr)$
      & $12418\text{K}$ \\
    \midrule
    \end{tabular}
    \end{adjustbox}
\end{table}

\subsection{Evaluation Metrics}
\label{subsec:measures}

In this work, we primarily focus on assessing the performance of the different baselines regarding their ability to detect \emph{when} an appliance was used and the underlying power estimated. 
We note that we also study the ability of CamAL to detect \emph{if} an appliance has been used in a given series using standard classification metrics.
In addition, as we proposed an ensemble approach, we evaluate the training time of our solution compared to the other baselines.
More precisely, we use the following measures:



\noindent \textbf{[Appliance Localization and Energy Estimation]}: 
The F1 Score, defined as the harmonic mean of Precision (Pr) and Recall (Rc), is used to evaluate the model’s predictive performance by balancing correct detections against false positives.
To measure the quality of energy estimation, we used the standard Mean Absolute Error (MAE) and Root Mean Square Error (RMSE).
In addition, we used the Matching Ratio (MR), based on the overlapping rate of true and estimated prediction, and considered as the best indicator performance for energy disaggregation~\cite{Mayhorn2016LoadDT}:
$MR = \frac{\sum_{t=1}^{T} min(\hat{y}_t, y_t)}{\sum_{t=1}^{T} max(\hat{y}_t, y_t)}$, where $T$ represents the total number of intervals, $y_t$ is the true and $\hat{y}_t$ is the predicted power usage of an appliance.

\noindent \textbf{[Appliance Detection]}: The F1 Score is widely used to benchmark binary classification problems with imbalanced data.
Nevertheless, this measure is often applied only to the minority class. 
However, in appliance detection scenarios, the minority class may vary, depending on the frequency of use of the appliance and the subsequence window length used for generating the dataset.
Therefore, to evaluate overall performance and account for variability, we used the Balanced Accuracy, that provides an indicator regardless of the minority class:
$\text{Balanced Accuracy} = \frac{1}{2} \left( \frac{TP}{TP + FN} + \frac{TN}{TN + FP} \right)$.

\noindent \textbf{[Scalability]}: 
We measured the training time (total and per epoch) as well as the inference time (throughput) to assess the scalability of the different approaches.
In particular, to evaluate the ability of our \emph{ensemble} solution to scale to large datasets of consumption series compared to other baselines.

Based on the experimental setup described above, we address the different Research Questions (RQs) enumerated in Section~\ref{sec:researchquestions} in the following sections.

\subsection{\textbf{RQ1:} Weakly vs Strongly Supervised Approaches}

In this Section, we answer RQ1 by comparing the performance of CamAL to other baselines and by evaluating the performance regarding the number of labels needed for training each method.
We evaluate each baseline by varying the number of instances (i.e., the number of subsequences) provided during the training. 
For the UKDALE dataset, which consists of only very few houses, we simply divide the dataset by percentage regardless of the number of houses.
For all the other datasets, we gradually add subsequence data from houses for training.
As a reminder, for the strongly supervised baselines, one subsequence corresponds to 510 labels (i.e., one label per timestamp).
In contrast, CamAL and the other weakly supervised baseline received one label per subsequence.



\subsubsection{Results}

\begin{figure*}
    \centering
    \includegraphics[width=1\linewidth]{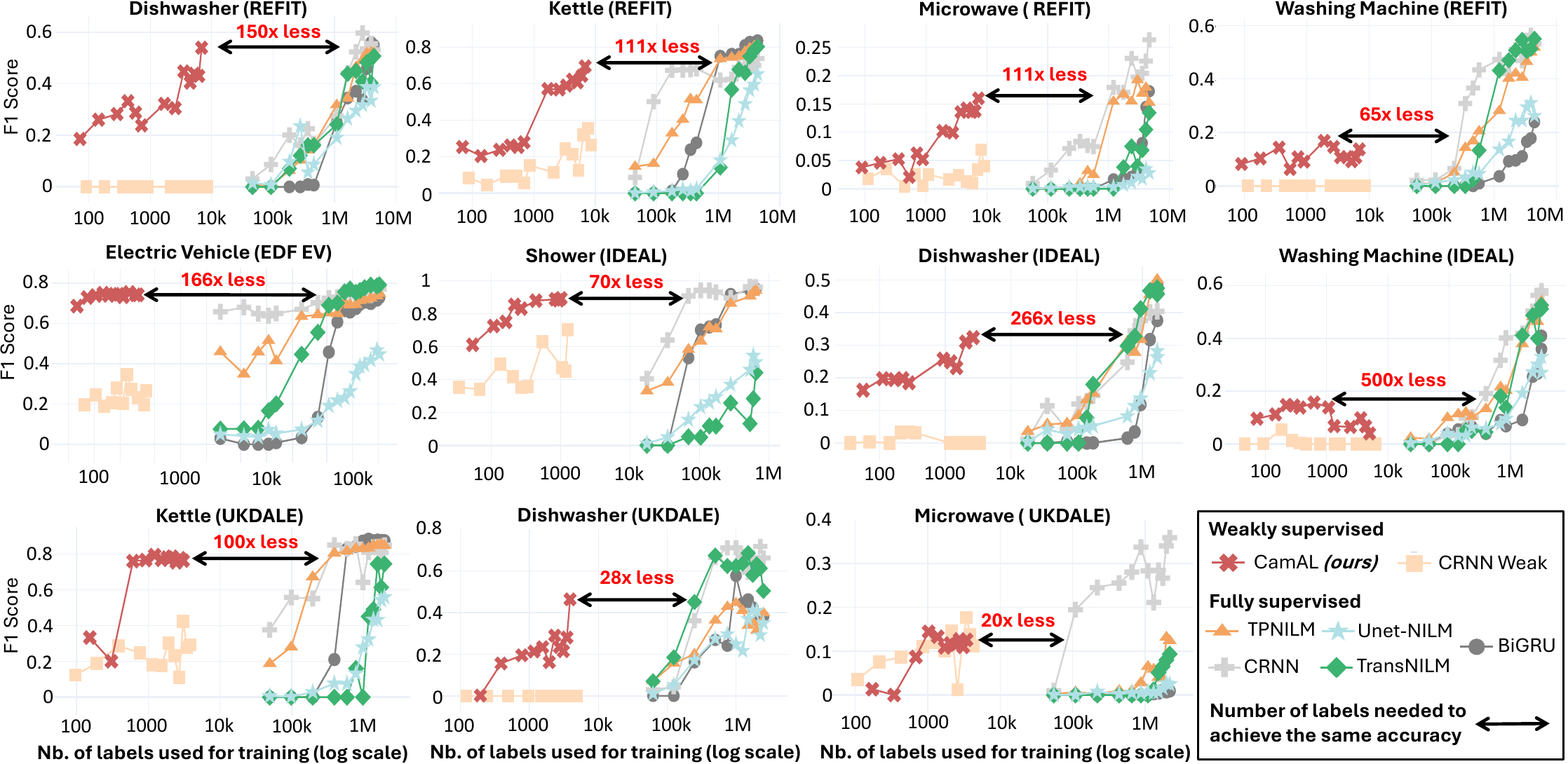}
    \caption{Overall results comparison according to the number of labels for training for each method.}
    \vspace{-0.3cm}
    \label{fig:resultscomparaison}
\end{figure*}

Figure~\ref{fig:resultscomparaison} reports the results for all the appliance localization of the 4 datasets for all the baselines in terms of accuracy (F1 Score) regarding the number of labels used for training each method.
The results show that for each case, the NILM baselines require significantly more labels to achieve the same accuracy as CamAL, from 20 times more for the Microwave case on the UKDALE dataset to 500 times more for the Washer case on the IDEAL dataset.
On average, we found that the NILM baselines require $144.27\times$ more labels to be able to achieve the same performance as CamAL.
However, we also note that in all the scenarios, the fully supervised baselines outperform our solution when using all the possible labels available at the cost of large differences in labels used for training.
Nevertheless, in 5 cases out of 11, CamAL almost equals the performances of strongly supervised approaches.
In addition, we note that CamAL significantly outperforms CRNN Weak regardless of the number of labels used for training for almost all datasets and appliances. 

\begin{table}
\caption{Weakly supervised approaches results (averaged over 5 runs).}
\label{table:results}
\begin{adjustbox}{width=\columnwidth, center}
\centering
\begin{tabular}{cc||cccc|cccc} 
\toprule

 & & \multicolumn{4}{c}{\textbf{CamAL}} & \multicolumn{4}{c}{\textbf{CRNN (Weak)}} \\

Datasets & Case & F1 & MAE & RMSE & MR & F1 & MAE & RMSE & MR \\

\midrule
\multirow{4}{*}{REFIT}  
& Dishwasher & \textbf{\textcolor{black}{0.54}} & \textbf{\textcolor{black}{44.8}} & \textbf{\textcolor{black}{242.3}} & \textbf{\textcolor{black}{0.2}} & 0.0 & 50.5 & 295.6 & 0.0 \\
& Kettle & \textbf{\textcolor{black}{0.7}} & \textbf{\textcolor{black}{10.8}} & \textbf{\textcolor{black}{125.6}} & \textbf{\textcolor{black}{0.48}} & 0.3 & 353.6 & 664.2 & 0.1 \\
& Microwave & \textbf{\textcolor{black}{0.16}} & \textbf{\textcolor{black}{4.3}} & \textbf{\textcolor{black}{64.6}} & \textbf{\textcolor{black}{0.09}} & 0.0 & 5.6 & 72.3 & 0.0 \\
& Washer & \textbf{\textcolor{black}{0.14}} & 19.6 & \textbf{\textcolor{black}{172.7}} & \textbf{\textcolor{black}{0.03}} & 0.0 & \textbf{\textcolor{black}{19.5}} & 176.3 & 0.0 \\

\midrule
\multirow{3}{*}{UKDALE}
& Dishwasher & \textbf{\textcolor{black}{0.46}} & 40.4 & 273.8 & \textbf{\textcolor{black}{0.03}} & 0.0 & \textbf{\textcolor{black}{35.5}} & \textbf{\textcolor{black}{252.4}} & 0.0 \\
& Kettle & \textbf{\textcolor{black}{0.76}} & \textbf{\textcolor{black}{20.9}} & \textbf{\textcolor{black}{191.7}} & \textbf{\textcolor{black}{0.3}} & 0.3 & 61.9 & 267.6 & 0.1 \\
& Microwave & \textbf{\textcolor{black}{0.13}} & \textbf{\textcolor{black}{6.9}} & 81.3 & 0.0 & 0.1 & 19.8 & \textbf{\textcolor{black}{77.1}} & \textbf{\textcolor{black}{0.03}} \\

\midrule
\multirow{3}{*}{IDEAL}
& Dishwasher & \textbf{\textcolor{black}{0.32}} & \textbf{\textcolor{black}{10.6}} & \textbf{\textcolor{black}{116.7}} & \textbf{\textcolor{black}{0.11}} & 0.0 & 10.8 & 125.9 & 0.0 \\
& Shower & \textbf{\textcolor{black}{0.89}} & \textbf{\textcolor{black}{9.7}} & \textbf{\textcolor{black}{131.8}} & \textbf{\textcolor{black}{0.8}} & 0.7 & 39.8 & 489.4 & 0.5 \\
& Washer & \textbf{\textcolor{black}{0.04}} & 14.7 & \textbf{\textcolor{black}{151.1}} & \textbf{\textcolor{black}{0.01}} & 0.0 & \textbf{\textcolor{black}{14.6}} & 151.8 & 0.0 \\

\midrule
EDF EV  
& EV & \textbf{\textcolor{black}{0.74}} & \textbf{\textcolor{black}{230.0}} & \textbf{\textcolor{black}{850.2}} & \textbf{\textcolor{black}{0.46}} & 0.3 & 2371.0 & 2818.9 & 0.1 \\

\midrule

\multicolumn{2}{c||}{\textbf{Avg.}} & \textbf{\textcolor{black}{0.38}} & \textbf{\textcolor{black}{38.5}} & \textbf{\textcolor{black}{227.2}} & \textbf{\textcolor{black}{0.23}} & 0.16 & 273.6 & 522.4 & 0.07 \\

\bottomrule

\end{tabular}
\end{adjustbox}
\end{table}

Table~\ref{table:results} reports the detailed results of our method compared to the other weakly supervised one (CRNN Weak) using all the instances (label) available for all the cases and datasets.
The results demonstrate that CamAL significantly outperforms the other weakly supervised methods on all the datasets and appliance localization scores. 
More specifically, we note an improvement of the average score on all datasets and cases of more than 135\% in terms of F1 Score and 247\% in MR.

The training time of the different baselines averaged across all cases, and datasets are reported in Figure~\ref{fig:scalability}(a) (left), while the training time according to the number of labels used for training for the IDEAL dataset (averaged for all cases) is shown in Figure~\ref{fig:scalability}(a) (right).
The results show that CamAL is among the two fastest solutions and is much faster than the other weakly supervised baseline, despite being an ensembling method. 
These results corroborate the theoretical analysis carried out in Section~\ref{sec:evalpipeline}.

\subsection{\textbf{RQ2}: Classification vs Localization}
\label{sec:rq2}

In this Section, we answer RQ2, which aims to understand the correlation between the detection (i.e., Problem~\ref{prob:detection} framed as a classification problem) performance of CamAL and its appliance localization pattern ability (i.e., Problem~\ref{prob:localisation}).

\begin{figure*}[tb]
    \centering
    \includegraphics[width=1\linewidth]{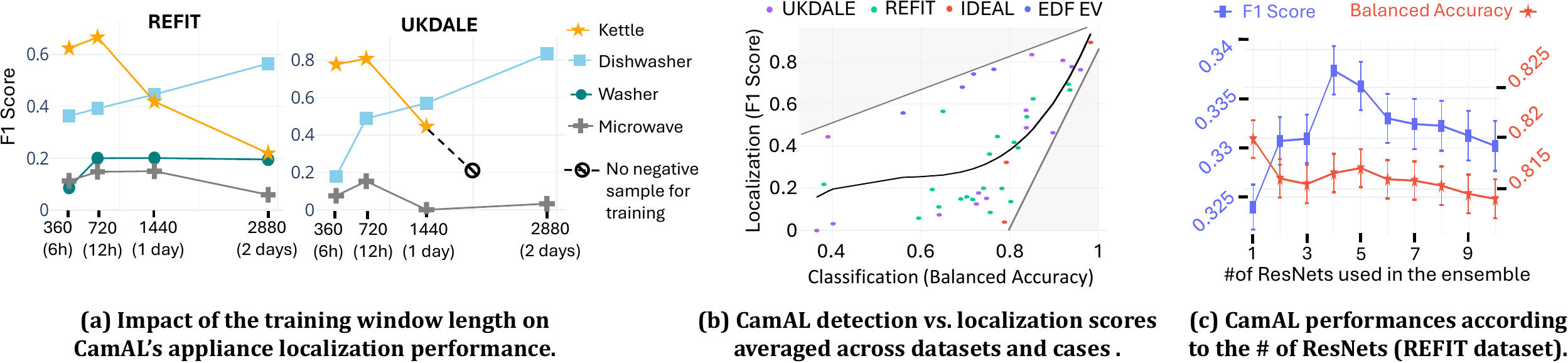}
    \vspace{-0.5cm}
    \caption{CamAL performance analysis: (a) effect of training window length, (b) detection vs. localization scores, and (c) influence of the number of ResNets.}
    \vspace{-0.4cm}
    \label{fig:camalperfanalysis}
\end{figure*}

Each point of the scatter plot on Figure~\ref{fig:camalperfanalysis}(b) corresponds to the performances obtained by CamAL for each appliance and dataset. 
The y-axis shows the localization score (F1 Score) according to the classification score (Balanced Accuracy) shown on the x-axis.
The results demonstrate a specific correlation between these two scores, highlighted by the 3rd-order regression line plotted on the graph.
More specifically, we note that reaching a good accuracy (more than 0.9) implies getting a good localization of the appliance pattern (more than 0.7 in terms of F1 Score).
However, the reciprocity is not true; for example, a relatively good localization performance is reached with a lower classification accuracy on the EDF EV dataset, while the same detection performance does not provide good localization performance in other cases on other datasets.
Nevertheless, the detection accuracy can be used as a proxy to assess the localization accuracy (especially in cases when the labels are not available).

\subsection{\textbf{RQ3}: Ablation Studies}
\label{sec:rq3}

\begin{table}[tb]
\caption{Influence of CamAL design choice on performance (for all REFIT appliances and averaged over 10 runs).}
\label{table:influencehparams}
\centering
\begin{tabular}{r|ccc}
\toprule
Metric & CamAL & w/o Attention module & w/o Different kernel $k_p$ \\
\midrule
F1 $\uparrow$ & \textbf{0.336} & 0.165 \textit{(-50.85\%)} & 0.317 \textit{(-5.6\%)} \\
Pr $\uparrow$ & \textbf{0.511} & 0.159 \textit{(-68.85\%)} & 0.499 \textit{(-2.21\%)} \\
Rc $\uparrow$ & 0.291 & \textbf{0.300} \textit{(+3.08\%)} & 0.275 \textit{(-5.68\%)} \\
MAE $\downarrow$ & \textbf{21.096} & 26.843 \textit{(-27.25\%)} & 21.336 \textit{(-1.14\%)} \\
MR $\uparrow$ & \textbf{0.162} & 0.114 \textit{(-29.59\%)} & 0.156 \textit{(-3.82\%)} \\
\bottomrule
\end{tabular}
\end{table}

In this Section, we perform different experiments to assess the influence of key parameters (answering RQ3).
First, we study the influence of window length on CamAL performances.
In other words, \textit{how weak can the labels be?}
Then, we conduct an ablation study to assess the proposed design choice of CamAL; we study the influence of (1) the number of ResNets used inside the CamAL ensemble, (2) the diversity of the size of kernels used in our networks, and (3) the importance of the attention-sigmoid module.

\subsubsection{How Weak Can the Labels Be?}

Figure~\ref{fig:camalperfanalysis}(a) shows the influence of the window length used for training CamAL on the localization performances reached for the different appliances using the UKDALE and REFIT datasets.
We note that to match with the rest of our experimental studies, the testing set remains the same, and we use subsequences of length 510.
The results demonstrate that for small appliances such as Kettle and Microwave, CamAL benefits from small windows (one label every 6 hours or 12 hours), while it is the opposite for big appliances, for which CamAL seems to perform better using longer windows. 
This can be explained by the fact that the Kettle and the Microwave are used mostly daily for these two datasets.
Therefore, using 1 or 2 days led to obtaining a really unbalanced dataset and, therefore, little data available for training our method after balancing the class.
This phenomenon is highlighted by the fact that it was not possible to train CamAL using a window length of 2 days on the UKDALE dataset, as no negative samples were available.
Consequently, the window length impacts the accuracy as it affects our ability to build a balanced training set. 
For large enough datasets with a balanced number of households with and without the corresponding appliances, the impact of the window length would be minimal.

\subsubsection{What Is the Impact of CamAL’s Design on Performance?}

Figure~\ref{fig:camalperfanalysis}(c) shows the results averaged over all the cases of the REFIT dataset in terms of localization (F1 Score) and classification score (Balanced Accuracy) by varying the number of ResNet $n$ used in the ensemble from 1 to 15.
We can notice that the classification score is stable regarding the number of ResNet used in the ensemble. 
In contrast, the localization score varies according to the number of classifiers used: it is minimal when using only one ResNet, reaches a peak around 4-5 ResNets, and then decreases. 
These results confirm that using an ensemble of ResNet instead of a single one leads to better localization performance.
However, using too many classifiers can slightly hurt CamAL performances.

Table~\ref{table:influencehparams} regroup the results of the ablation studies conducted on the REFIT dataset (averaged over all the cases) to study the influence of CamAL design.
The first column shows the results for CamAL; the second shows the results for CamAL ablated from the Attention-Sigmoid module, and the last columns correspond to CamAL using an ensemble of ResNets that doesn't use different kernel size (we set $k_p=7$ for all the ResNets, as originally proposed in~\cite{ResNet_tsc_2017}).
First, we notice that using the Attention-Sigmoid module greatly improves the overall performances of CamAL by more than 50\% in terms of F1 Score and nearly 70\% in terms of Precision.
This highlights that only using the Average of the CAM extracted from the ensemble doesn't suit our problem.
In fact, using the raw average of the CAMs leads to obtaining a slightly better Recall while all the other metrics are negatively impacted, meaning that the number of false activations is too high.
Secondly, the results obtained using a fixed kernel in each ResNet lead to a slight drop in the results, demonstrating the importance of using different receptive fields to obtain different activation maps.

\begin{figure*}[tb]
    \centering
    \includegraphics[width=1\linewidth]{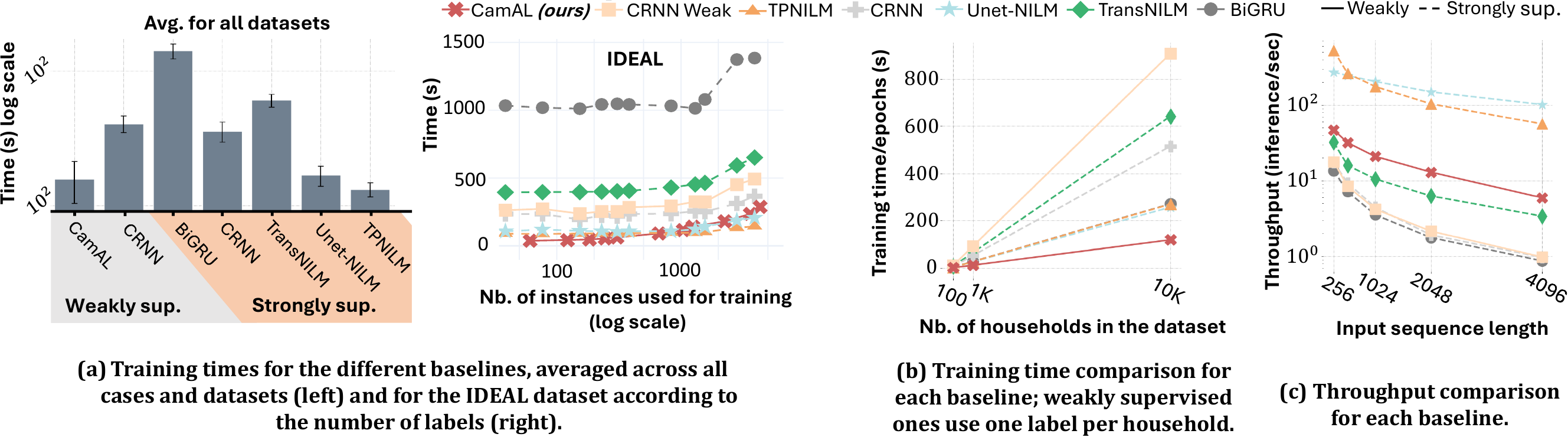}
    \vspace{-0.5cm}
    \caption{Comparisons of (a) average training time; (b) per epoch training time when varying the \# of households; and (c) running inference time.}
    \vspace{-0.3cm}
    \label{fig:scalability}
\end{figure*}

\subsection{\textbf{RQ4}: An Extreme (Yet Realistic) Scenario with Only One Weak Label per Household}
\label{sec:application}

\begin{figure}[tb]
    \centering
    \includegraphics[width=1\linewidth]{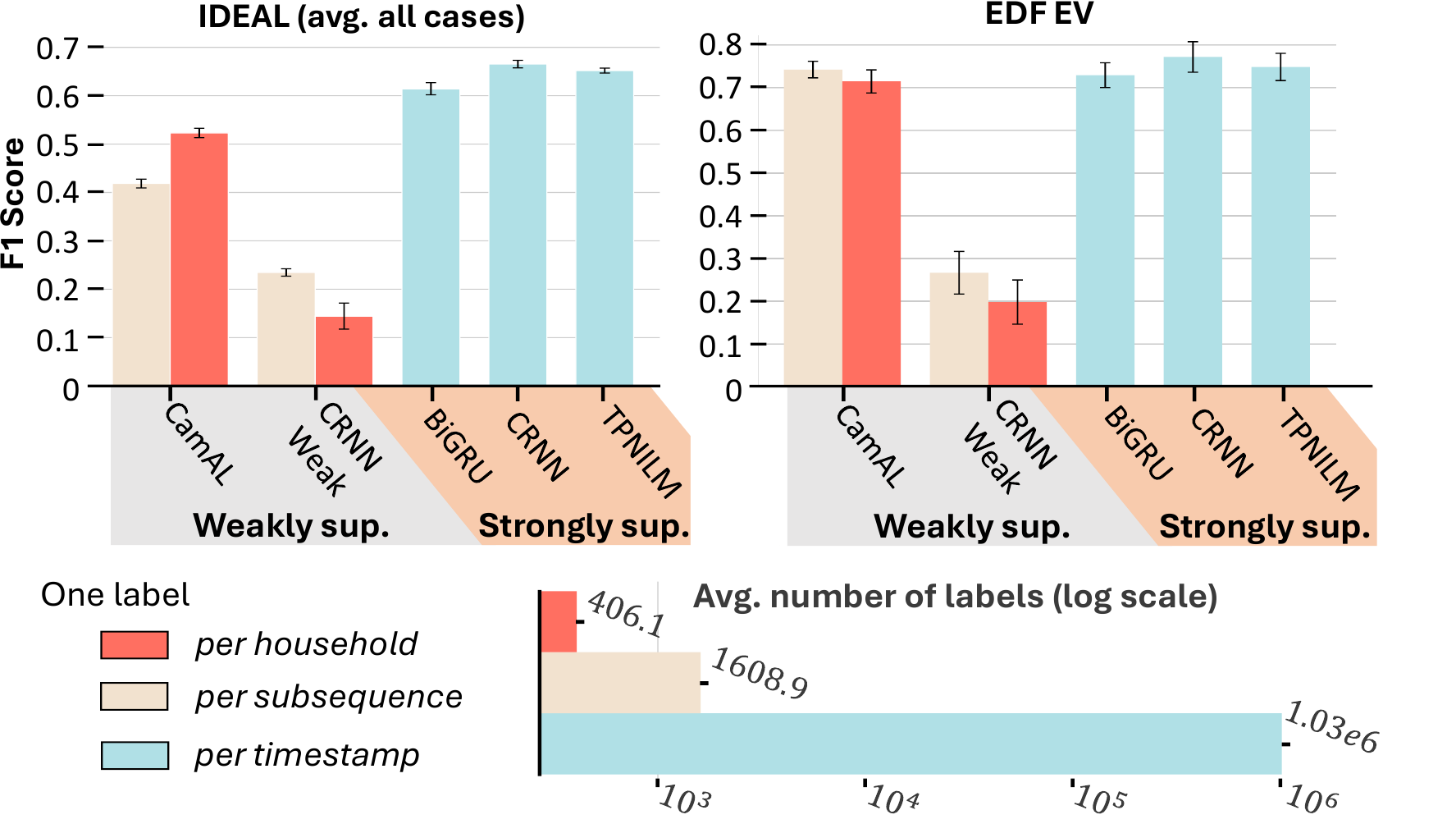}
    \vspace{-0.6cm}
    \caption{Results comparison for the baselines trained using the different types of labels (one label per household, per subsequence, or per timestamp).}
    \vspace{-0.cm}
    \label{fig:resultspossession}
\end{figure}

\begin{figure}[tb]
    \centering
    \includegraphics[width=1\linewidth]{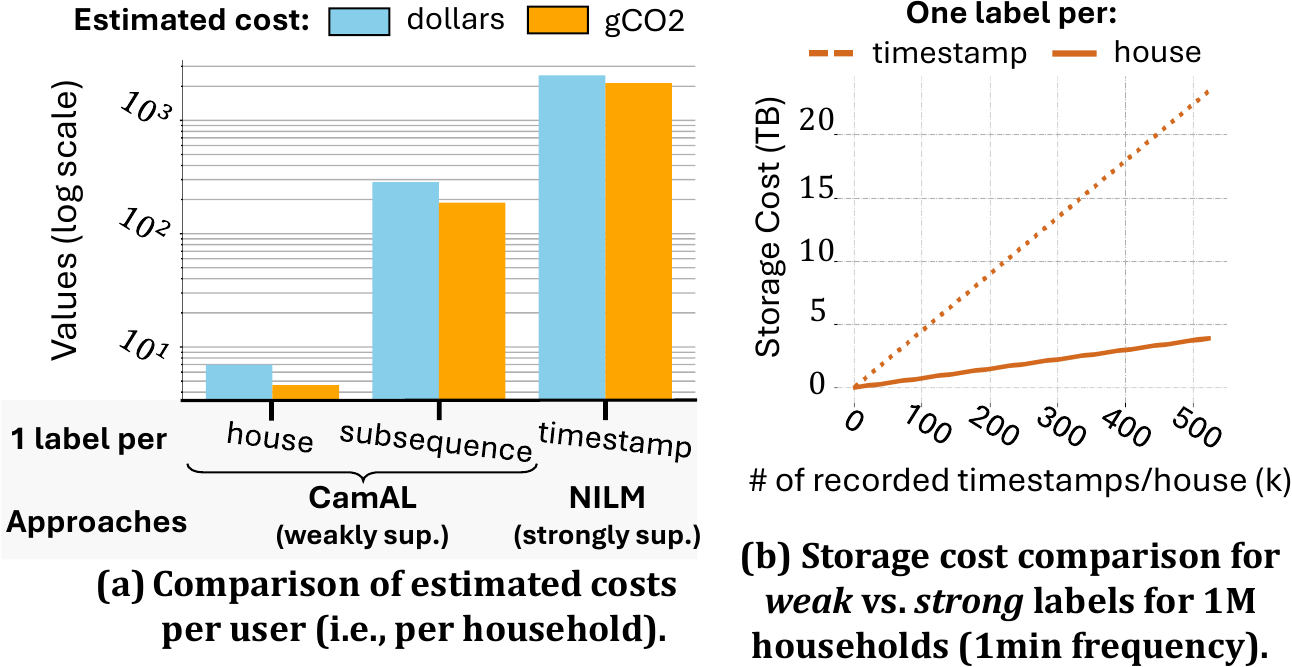}
    \vspace{-0.5cm}
    \caption{Costs comparison: (a) dollars and $gCO_2$ per user; (b) Storage.}
    \vspace{-0.5cm}
    \label{fig:costs}
\end{figure}

Although we demonstrate in the previous sections the benefits of CamAL regarding the low number of labels needed to achieve similar performance as NILM approaches, the training still relies on expensive to gather datasets (one label per time interval).
However, one question that arises is: \textit{Can we actually use only one label?} (i.e., one label indicating that the user has the appliance or not). 
This question is particularly important because, if successful, CamAL would require the practitioner to ask only once about the list of appliances that their consumers have in their household. 

We study this extreme (but realistic) scenario by evaluating the accuracy of our approach, as well as the implications in terms of monetary cost and carbon footprint.


\subsubsection{Performance Comparison of the Different Approaches}

We performed experiments using the IDEAL dataset, the EDF EV, and the EDF Weak datasets (which are the only two datasets that provide enough possession-level information). 
For the IDEAL datasets, we used the possession information of the different appliances owned by the 255 households provided in the questionnaires to get the label for our training dataset. 
Subsequently, the per-timestamp labeled subgroup of 39 households was used to test the method's performance on ground truth data.
For EDF datasets, we use the EDF Weak dataset as training, while the EDF EV dataset (which is labeled per timestamp) is for the test.

\noindent{\textbf{[Possession Only Pipeline]}} The two weakly labeled datasets used for training (the 255 households of the IDEAL dataset and the EDF Weak dataset) are composed of variable length electricity consumption series 
and \emph{labels of possession} for different appliance $a$. 
We divide the datasets in a standard 70\%/10\%/20\% random split for the training, validation, and test sets.
We first balanced the training set through random undersampling to equalize class distribution. 
Following prior work~\cite{VLDB_TransApp}, we then sliced household consumption into smaller subsequences to augment the training data, experimenting with various tumbling window sizes $w$: 
for IDEAL, we tested $w = \{1440, 2880, 5760, 10080, 20160, 30240, 40320, 50400\}$, corresponding to time windows ranging from one day to five weeks; 
for EDF Weak we used $w = \{256, 512, 1024\}$.
Note that the label of the entire consumption series (i.e., \emph{label of possession}) is assigned to all sliced subsequences during the training process without any other information.
We then test the baselines using the same setting 
as reported in Section~\ref{sec:evalpipeline}.
Note that the reported localization score corresponds to the best score reached in terms of classification (Balanced Accuracy) for a given window length $w$.


\noindent{\textbf{[Results]}} The results are reported in Figure~\ref{fig:resultspossession}.
For comparisons, we report the accuracy of strongly supervised methods and weakly supervised approaches trained with labels per subsequences obtained in the previous sections.
On the IDEAL dataset, CamAL trained on household possession labels achieves better results than when trained using subsequences from the 39 submetered households.
Moreover, as shown in Figure~\ref{fig:introfig}, CamAL uses more than 5200 times fewer labels than strongly supervised methods while achieving nearly the same accuracy in the Dishwasher scenario of the IDEAL dataset.
Finally, experiments on the EDF datasets demonstrate that training with possession information yields results equivalent to  CamAL trained with one label per subsequence and comparable to those obtained by strongly supervised methods. 

Interestingly, the CRNN baseline performs worse on both datasets when trained with possession labels than when trained with labels per subsequence.
Overall, these results demonstrate that training CamAL on the appliance detection task and using only the possession label can enable good localization results.

\subsubsection{Cost Comparison of the Different Approaches}

As mentioned earlier, building NILM datasets can be costly, which is the main motivation for proposing methods that can operate with weak labels.
In this section, we compare the cost of collecting and storing real submeter appliance consumption data (typically NILM datasets) with surveys that ask consumers to complete a questionnaire (such as EDF Weak).

EDF must invest approximately \$1000/household in sensors and another \$1500/household per year in maintenance to collect different appliances' submeter signals. 
Gathering the possession information of each appliance owned in the household is done by sending a simple questionnaire that the customers fill out for a total cost of \$10/household.
As an electricity supplier that wants to achieve net zero carbon in France, EDF is also concerned about the $CO_2$ emission of such a deployed solution.
To monitor a household, the company has to send a technician to instrument the house with sensors for an average $CO_2$ emission cost of at least $2134g$ (assuming car $CO_2$ emission of $97g/km$ and an average commute distance in France of $22km$~\cite{ADEME_CO2_Emissions, ObservatoireDesTerritoires_MobilitesQuotidiennes}).
On the other hand, a study recently estimates the cost of visiting a website to be around $4.62gCO_2$~\cite{RESET_Carbon_Footprint_Website}, which can be seen as a lower bound of a dedicated website built for consumers to answer a questionnaire on the appliances in their household.

Consider also that each timestamp of recorded electricity consumption data is stored in BIGINT values (8 bytes), while the possession information is stored in VARCHAR values (10 bytes).
The costs of these two solutions are reported in Figure~\ref{fig:costs}.
Figure~\ref{fig:costs}(a) shows that obtaining the label to be able to train the supervised method is by far the most expensive in terms of both money (\$) and emissions ($gCO2$).
Conversely, asking consumers to answer surveys (daily or weekly) to obtain labels on subsequences reduces both costs by an order of magnitude, and asking for the possession information only (what CamAL uses), further reduces both costs by more than an order of magnitude. 
Moreover, Figure~\ref{fig:costs}(b) shows that collecting strong labels for 1 million households for 5 appliances every minute results in $\sim$15TB/year, 6x more data than simply collecting weak labels (appliance possession information only). Overall, when compared to the strong label needs of current NILM solutions, gathering the weak labels to train CamAL reduces the monetary cost and carbon footprint by $>$2 orders of magnitude while also drastically reducing the storage cost, leading to a truly scalable solution.

\subsubsection{Assessing CamAL's Scalability}

To assess the baseline's real-world scalabilities to larger datasets, we performed experiments on synthetic data to measure the training time per epoch according to the number of households.
More specifically, we generated a random consumption dataset (i.e., white noise), including both total aggregated consumption and per-timestamp appliance ground truth labels) at a 30-minute sampling rate (i.e., series of length 17520). 
Indeed, sequence-to-sequence NILM approaches need to operate on subsequences of an entire consumption series of a household to be trained to achieve suitable performances (e.g., windows of length 510)~\cite{ontheimpactofwinlengthNILM}. 
We trained all the baselines on a single GPU using a batch size of 64.
For all the strongly supervised baselines, the entire sequences are first broken down into smaller subsequences of length 510; in contrast, the two weakly supervised approaches take the whole sequence directly as input. 
As shown in Figure~\ref{fig:scalability}(b), CamAL remains substantially more efficient than strongly supervised NILM baselines according to the number of households used for training, demonstrating the potential real-world aspect of our approach when applied to large-scale datasets.

Figure~\ref{fig:scalability}(c) shows the throughput (inference/sec) by varying the input subsequence length given as input, measured on a single CPU.
First, we can see that CamAL is significantly more efficient than the other weakly supervised baseline (CRNN Weak). 
In addition, we note that CamAL is more efficient than three out of the 5 NILM baselines. 
The only two more efficient are convolutional-based baselines (TPNILM and Unet-NILM), but they require far more labels to be trained. 

\subsection{\textbf{RQ5}: A Data Augmentation Perspective}
\label{sec:augmentation}

\begin{figure}[tb]
    \centering
    \includegraphics[width=0.95\linewidth]{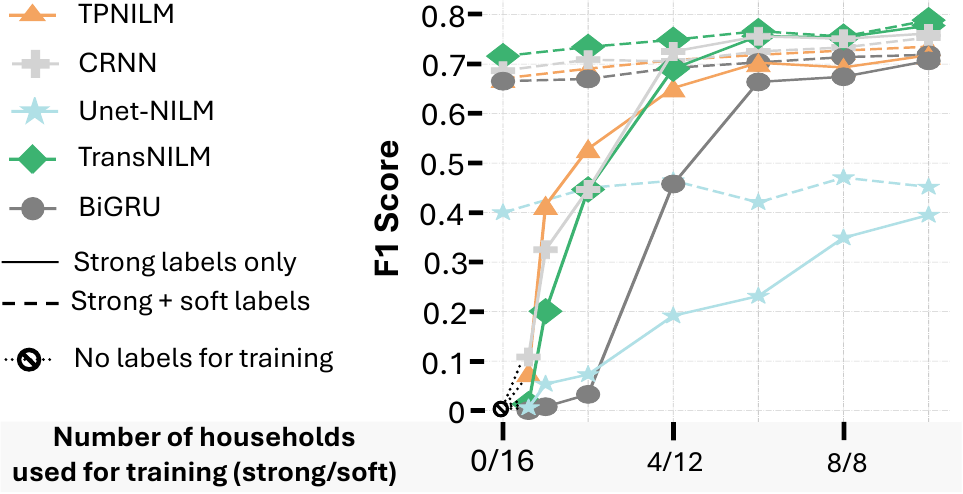}
    \vspace{-0.1cm}
    \caption{Performance of strongly supervised baselines trained on CamAL soft labels (i.e., outputs) using the EDF EV dataset.}
    \vspace{-0.3cm}
    \label{fig:softlabelcamal}
\end{figure}

As a perspective of our proposed approach, we investigate in this final Section the use of our method for generating \emph{soft labels} that can be used to enhance the performance of strongly supervised NILM approaches in case of lack of strong labels.
Trained on the EDF Weak dataset for EV detection (Sec.~\ref{sec:application}), CamAL outputs are used as soft labels for the EDF EV dataset. 
In the most extreme case, no ground truth data are used, only CamAL's predictions. 
Subsequently, we incrementally add ground truth labels from an increasing number of houses (up to 8) to assess performance improvements.

The results shown in Figure~\ref{fig:softlabelcamal} demonstrate that the supervised baselines can be trained using only soft labels without a significant loss in accuracy. 
Additionally, when ground truth labels are scarce, all baselines achieve significantly better results by combining both strong and soft labels.
For example, when using strong labels in at most one household, adding soft labels improves results between 34\% (for TPNILM) and 1200\% (for BiGRU).
These results open new directions, including improving CamAL soft labels to obtain individual appliance power.
While multiplying the localized binary signal by a single average power rating is a useful simplification, more advanced post-processing methods are needed to refine the estimated consumption further. 

\section{Conclusions}
We introduced CamAL, a weakly supervised approach for appliance pattern localization that only requires knowing the presence (or not) of the appliance in a household. 
By leveraging an ensemble of deep-learning classifiers combined with explainable classification methods, 
CamAL significantly reduces the need for strong per-timestamp labels and can be trained using only appliance possession information. 
Our experiments on 4 real-world datasets have shown that CamAL not only outperforms existing weakly supervised baselines but also reaches comparable performance to fully supervised NILM approaches while using considerably fewer labels.
This makes CamAL the first truly non-invasive solution for load monitoring, aligning well with the needs of electricity suppliers and households seeking to avoid unnecessary installation costs and carbon emissions.

CamAL is already in use within EDF for internal consumption analyses and also available online as a demo~\cite{2025_petralia_devicescope_icde}, and plans for broader deployment are under consideration. 
Overall, CamAL opens a new direction in NILM research, proving that effective appliance localization can be achieved with minimal supervision using explainability-based approaches.

\section*{Acknowledgments}

Supported by EDF R\&D, ANRT French program, and EU Horizon projects AI4Europe (101070000), TwinODIS (101160009), ARMADA (101168951), DataGEMS (101188416), RECITALS (101168490).

\bibliographystyle{IEEEtran}
\bibliography{bibliography}

\end{document}